\documentclass{sig-alternate-2013}

\usepackage[
  pass,
]{geometry}

\newfont{\mycrnotice}{ptmr8t at 7pt}
\newfont{\myconfname}{ptmri8t at 7pt}
\let\crnotice\mycrnotice%
\let\confname\myconfname%

\usepackage{url}
\usepackage{graphicx}
\usepackage{caption}
\usepackage{subcaption}

\begin{document}     
%

\title{ALOJA-ML: A Framework for Automating Characterization and Knowledge Discovery in Hadoop Deployments}

\numberofauthors{6}
\author{
\alignauthor
Josep Ll. Berral\\
       \affaddr{Barcelona Supercomputing Center (BSC)}\\
       \affaddr{Universitat Polit{\`e}cnica de Catalunya (UPC)}\\
       \affaddr{Barcelona, Spain}
\alignauthor
Nicolas Poggi\\
       \affaddr{Barcelona Supercomputing Center (BSC)}\\
       \affaddr{Universitat Polit{\`e}cnica de Catalunya (UPC)}\\
       \affaddr{Barcelona, Spain}
\alignauthor
David Carrera\\
       \affaddr{Barcelona Supercomputing Center (BSC)}\\
       \affaddr{Universitat Polit{\`e}cnica de Catalunya (UPC)}\\
       \affaddr{Barcelona, Spain}
\and  
\alignauthor
Aaron Call\\
       \affaddr{Barcelona Supercomputing Center (BSC)}\\
       \affaddr{Barcelona, Spain}
\alignauthor
Rob Reinauer\\
       \affaddr{Microsoft Corporation}\\
       \affaddr{Redmond, USA}
\alignauthor
Daron Green\\
       \affaddr{Microsoft Corporation}\\
       \affaddr{Microsoft Research (MSR)}\\
       \affaddr{Redmond, USA}
}

\maketitle
\begin{abstract}
This article presents ALOJA-Machine Learning (ALOJA-ML) an extension to the ALOJA project that uses machine learning techniques to interpret Hadoop benchmark performance data and performance tuning; here we detail the approach, efficacy of the model and initial results.
The ALOJA-ML project is the latest phase of a long-term collaboration between BSC and Microsoft, to automate the characterization of cost-effectiveness on Big Data deployments, focusing on Hadoop. 
Hadoop presents a complex execution environment, where costs and performance depends on a large number of software (SW) configurations and on multiple hardware (HW) deployment choices. 
Recently the ALOJA project presented an open, vendor-neutral repository, featuring over 16.000 Hadoop executions. These results are accompanied by  a test bed and tools to deploy and evaluate the cost-effectiveness of the different hardware configurations, parameter tunings, and Cloud services\footnote{ALOJA's repository and sources at http://hadoop.bsc.es/}.

Despite early success within ALOJA from expert-guided benchmarking~\cite{DBLP:conf/bigdataconf/PoggiCCMBTAGLRVGB14}, it became clear that a genuinely comprehensive study requires automation of modeling procedures to allow a systematic analysis of large and resource-constrained search spaces.
ALOJA-ML provides such an automated system allowing knowledge discovery  by modeling Hadoop executions from observed benchmarks across a broad set of configuration parameters. 
The resulting empirically-derived performance models can be used to forecast execution behavior of various workloads; they allow  {\em a-priori} prediction of the execution times for new configurations and HW choices and they offer a route to model-based anomaly detection.  In addition, these models can guide the benchmarking exploration efficiently, by automatically prioritizing candidate future benchmark tests. 
Insights from ALOJA-ML's models can be used to reduce the operational time on clusters, speed-up the data acquisition and knowledge discovery process, and importantly, reduce running costs.
In addition to learning from the methodology presented in this work, the community can benefit in general from ALOJA data-sets, framework, and derived insights to improve the design and deployment of Big Data applications.

\end{abstract}

\category{C.4}{Performance of Systems}{Modeling techniques}
\category{D.4.8}{Operating Systems}{Performance}[Modeling and Prediction]
\category{I.2.6}{Artificial Intelligence}{Learning}[Induction,Knowledge acquisition]

\terms{Experimentation; Measurement; Performance}

\keywords{Data-Center Management; Modeling and Prediction; Machine Learning; Execution Experiences; Hadoop}

\section{Introduction} 

Hadoop has emerged as the {\em de-facto} framework for Big Data processing deployment~\cite{apache:hadoop}\cite{White:2009:HDG:1717298} and its adoption continues at a compound annual growth rate of 58\% \cite{person14}. 
Even with this impressive trend, deploying and running a cost-effective Hadoop cluster is  hampered by the extreme complexity of its distributed run-time environment and the large number of potential deployment choices. 
It transpires that many of the software parameters exposed by both Hadoop and the Java run-time  have quite a pronounced impact on job performance\cite{heger:hadoopapproach, heger:hadooptuning, Herodotou11starfish:a} and therefore a corresponding effect on the cost of execution~\cite{DBLP:conf/bigdataconf/PoggiCCMBTAGLRVGB14}.
Selecting the most appropriate deployment pattern for a given workload,  whether for on-premise servers or as part of a cloud service, involves a complex set of decisions.  Any {\em a-priori}  information presented as either heuristic or as a specific performance prediction could greatly improve the decision making process resulting in improved execution times and reduced costs.

\begin{figure} 
\centering
 \includegraphics[width=0.5\textwidth]{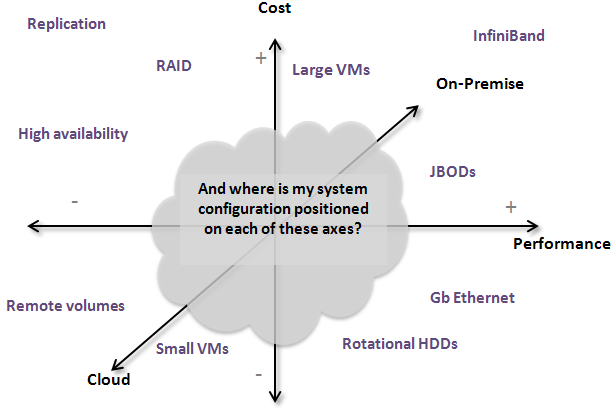}
\caption{A cloud of points for Cost vs. Performance vs. Cloud or On-premise}
\label{fig:cloud_vs_local}
\end{figure}

Assisting such decision making often requires manual and time-consuming benchmarking followed by operational fine-tuning for which few organizations have either the time or performance profiling expertise.
A problem inherent in such complex systems is the difficulty of finding a generalizable {\em rule of the thumb} for configurations that can be applied to all workloads (Hadoop jobs).
To illustrate that complexity, figure~\ref{fig:cloud_vs_local} presents an example of the search space for evaluating the cost-effectiveness of a particular workload and setup.  

ALOJA-ML provides  tools to automate both the knowledge discovery process and performance prediction of Hadoop benchmark data. It is the latest phase of the ALOJA initiative which is an on-going collaborative engagement between the Barcelona Supercomputing Center (BSC), Microsoft Product groups and Microsoft Research (MSR). The collaboration explores upcoming hardware architectures and builds automated mechanisms for deploying cost-effective Hadoop clusters. 
ALOJA-ML's machine-learning derived performance models predict execution times for given a workload and configuration (HW and SW) and provide configuration recommendations for optimal performance of a given task. Given the ability to forecast task performance, the tools can also be used to detect anomalous Hadoop execution behavior. This is achieved using a combination of machine learning modeling and prediction techniques, clustering methods, and model-based outlier detection algorithms.

\subsection{Motivation}
Dealing with Hadoop optimization necessarily includes running multiple executions and examining large amounts of data, from the information of the environment and configurations, through to the outputs from result logs plus performance readings. In this project we deal with all available data with data mining and machine learning techniques. Selecting which are the relevant features and/or discarding useless information is key to discovering which parameters have the biggest impact on performance, or even the opposite, which features can be freely modified without any detrimental performance impact, allowing the user/operator to adjust configurations not only to me specific performance needs but also to the available resources. Recommending a specific system configuration requires knowledge on how parameter selection will affect various potential goals (e.g., increased performance, reduced execution cost or even energy savings). Modeling and automated learning techniques provide not only the ability to predict the effects of a given configuration, but also they allow exploration of possible configurations {\em in-vitro}, characterizing how hypothetical scenarios would work. The goal being to demystify the black-art of Hadoop performance tuning by providing Hadoop users and operators with the ability to predict workload performance and empower them with a clear understanding of the effect of configuration selections and/or modifications in both the HW and SW stack.

\subsection{Contribution}

In ALOJA-ML we aim to provide 1) a useful framework for Hadoop users and researchers characterize and address configuration and performance issues; 2) data-sets of Hadoop experimentation and prediction; and 3) a generalized methodology and examples for automated benchmarking couple to data mining techniques.  This is achieved as follows:

\begin{enumerate}

\item The ALOJA framework for Big Data benchmarking, including the machine learning enhancements, is available to all the community to be used with own Hadoop data-set executions. Researchers or developers can implement or expand this tool-set through comparing or predicting data from observed task executions and/or by adding new machine learning or new anomaly detection algorithms.

\item All the data-sets collected for ALOJA and ALOJA-ML are public, and can be explored through our  framework or used as data-sets in other data analysis platforms.

\item In this work we share our experiences in analyzing Hadoop execution data-sets. We present results on modeling and prediction of execution times for a range of systems and Hadoop configurations. We show the results from our anomalous execution detection mechanisms and model-based methods for ranking configurations depending on relevance.  These rankings are used to characterize the minimal set of executions required to model a new infrastructure.

\end{enumerate}

This article is structured as follows: Section~\ref{sec:background} presents the preliminaries for this work and current state-of-art. 
Section~\ref{sec:prediction}, presents the datasets, the methodology and results for the Hadoop job execution time predictor model. 
Section~\ref{sec:experiences} presents two use cases and extension for the predictor model: execution anomaly detection and guided benchmarking, our approach to guide benchmarks executions by predictive value.  Finally, sections~\ref{sec:conclusions} and ~\ref{sec:future} the conclusions and future work lines for the project.   

\section{Background}
\label{sec:background}

This section presents the background to the project and the current state-of-the art in Hadoop and system performance modeling applying machine learning.

\subsection{The ALOJA project}
The work presented here is part of the ALOJA Project; an initiative of the Barcelona Supercomputing Center (BSC) that has developed Hadoop-related computing expertise for over 7 years~\cite{bsc:autonomic}.  The efforts are partially supported the Microsoft Corporation, that contributes technically through product teams, financially, and by providing resources as part of its global {\em Azure4Research} program.
ALOJA's initial approach was to create the most comprehensive vendor-neutral open public Hadoop benchmarking repository. It currently features over 16.000 Hadoop benchmark executions, used as the main data-set for this work.  The tools on the online platform currently compares not only software configuration parameters, but also \emph{current}, to new available hardware including SSDs, InfiniBand networks, and Cloud services. 
We also capture the cost of each possible setup along with the run time performance, with a view to offer configuration recommendations for a given workload.
As few organizations have the time or performance profiling expertise,   
we expect our growing repository and analytic tools  will benefit Hadoop community to meet their Big Data application needs.  The next subsection briefly presets the main platform components, as a more complete description of the scope and goals, the architecture of the underlying framework and the initial findings are presented in~\cite{DBLP:conf/bigdataconf/PoggiCCMBTAGLRVGB14}.

\subsubsection{Benchmarking}
\label{sec:benchmarking}

Due to the large number of configuration options that have an effect on Hadoop's performance, it has been necessary previous to this work to characterize Hadoop using extensive benchmarking.
Hadoop's distribution includes jobs that can be used to benchmark its performance, usually referred as \emph{micro benchmarks}, however these type of benchmarks usually have limitation on their representativeness and variety.
ALOJA currently features the HiBench open-source benchmark from Intel~\cite{HiBench}, which can be more realistic and comprehensive than the supplied example jobs in Hadoop. For a complete description please refer to~\cite{HiBench}, and a  characterization of the performance details for the benchmarks can be obtained in the \emph{Performance Charts} section of ALOJA's online application~\cite{bsc:hadoop}.

\subsubsection{Current Platform and Tools}

\begin{figure} 
\centering
 \includegraphics[width=0.5\textwidth]{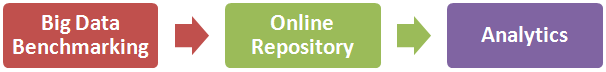}
\caption{Main components and workflow in the ALOJA framework}
\label{fig:components}
\end{figure} 

The ALOJA platform\footnote{ALOJA code available at: https://github.com/Aloja/}, is composed of open-source components to achieve an automated benchmarking of Big Data deployments, either on-premise or in the cloud.  To achieve this goal, clusters and nodes can be easily defined within the framework, and a set of specialized deployment and cluster orchestration scripts take care of the cluster setup.  Then, a set of benchmarks to explore and execute can be selected and queued for execution.  As benchmarks execute, results are gathered and imported into the main application ---that features a Web interface, feeding the repository.  On the Web-based repository of benchmark executions, the user can select to search and filter interesting executions and browse the execution details.  Advanced analytic features are also provided to perform normalized and aggregated evaluations of up to thousands of results.  The tools include:  

\begin{itemize}
  \item Best configuration recommendation for a given SW/HW combination.
  \item Speed up comparison of SW/HW combinations.
  \item Configuration parameter evaluation
  \item Cost/Performance analysis of single and clustered executions
  \item Cost-effectiveness of cluster setups
\end{itemize}  

Figure~\ref{fig:components} presents the main components of the platform that feed each other in a continuous loop. It is the intent of this work to add data-mining capabilities to the current online tools, to enable automated knowledge discovery and characterization, that is presented in later sections.

The platform also includes a {\em vagrant} virtual machine with a complete sand-box environment and sample executions that is used for development and early experimentation.  In our project's site~\cite{bsc:hadoop} there is further explanation and documentation of the developer tools.

\subsection{Related Work}

The emergence and the adoption of Hadoop by the industry has led to various attempts at performance tuning an optimization, schemes for data distribution or partition and/or adjustments in HW configurations to increase scalability or reduce running costs. For most of the deployments, execution performance can be improved at least by 3x from the default configuration~\cite{heger:hadooptuning,heger:hadoopapproach}. A significant challenge remains: to characterize these deployments and performance, looking for the optimal configuration in each case.
There is also evidence that Hadoop performs poorly with newer and scale-up hardware~\cite{Appuswamy2013}.
Scaling out in number of servers can usually improve performance, but at increased cost, power and space usage~\cite{Appuswamy2013}. 
These situations and available services make a case to reconsider scale-up hardware and new Cloud services from both a research and an industry perspective.

Previous research focused on the need for tuning Hadoop configurations to match specific workload requirements; for example, the Starfish Project from H. Herodotou~\cite{Herodotou11starfish:a} proposed to observe Hadoop execution behaviors and use profiles to recommend configurations for similar workload types. This approach is a useful reference for ALOJA-ML when modeling Hadoop behaviors from observed executions, in contrast, we have sought to use machine learning methods to  characterize the execution behavior across a large corpus of profiling data.

Some approaches on autonomic computing already tackled the idea of using machine learning for modeling system behavior vs. hardware or software configuration e.g., works on self-configuration like J.Wildstrom~\cite{Wildstrom:2007:MLO:1625275.1625456} used machine learning for hardware reconfiguration on large data-center systems. Similarly, P.Shivam' NIMO framework~\cite{Shivam:2006:AAL:1182635.11641734} modeled computational-science applications allowing prediction of their execution time in grid infrastructures. Such efforts are precedents of successful applications of machine learning modeling and prediction in distributed systems workload management. Here we apply such methodologies, not to directly manage the system but rather to allow users, engineers and operators to learn about their workloads in a distributed Hadoop environment.

\section{Modeling Hadoop with Machine Learning}
\label{sec:prediction}
The primary contribution of this work is the inclusion of data-mining techniques in the analysis of Hadoop performance data. Modeling Hadoop execution allows prediction of execution output values (e.g.,  execution time or resource consumption) based on input information such as software and hardware configuration. Modeling the system also enables anomaly detection by comparing actual executions against predicted outputs, flagging as anomalous those tasks whose run-time lies notably outside a machine-learned prediction. Furthermore, compressing system observations using clustering techniques identifies the set of points required for minimal characterization of the system, indicating the most representative tasks needed to model the system or possible sets of tasks needed to learn the behavior of a new system.

\begin{figure}[h!tbp]
\centering
\includegraphics[width=1.0\linewidth]{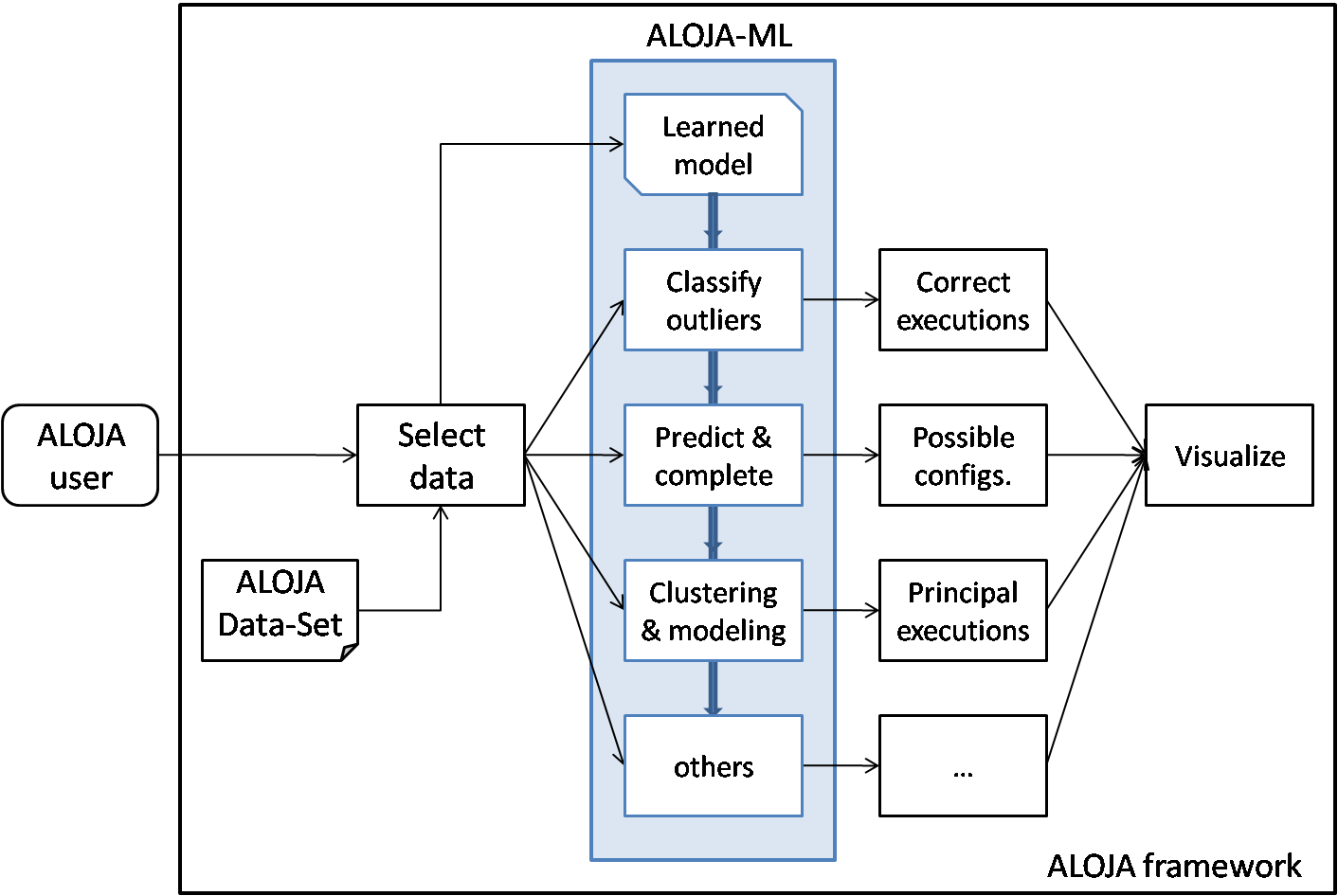}
\caption{ALOJA-ML inside the ALOJA Framework}
\label{figure:alojaml-schema}
\end{figure}

The ALOJA framework collates and analyzes data collected from Hadoop task executions and displays it through a range of tools; this helps users understand and interpret the executed tasks. ALOJA-ML complements this framework by adding tools that learn from the data and extract hidden (or not so obvious) information, also adding an intermediate layer of data treatment to complement the other visualization tools. Figure~\ref{figure:alojaml-schema} shows the role of ALOJA-ML inside the ALOJA framework.

\subsection{Data-sets and Structure}

The ALOJA data-set is an open-access collection of Hadoop traces, containing currently over 16.000 Hadoop executions of 8 different benchmarks from the Intel HiBench suite~\cite{HiBench}.  Each execution is composed of a prepare Hadoop job that generates the data e.g., \emph{teragen}, and a proper benchmark e.g., \emph{terasort}.  Although the job of interest is generally the proper benchmark (\emph{terasort}), prepare jobs are also valid jobs that can be also used for training models.  This leaves us with over 32.000 executions to learn from.
Each benchmark is run with different configurations, including clusters and VMs, networks, storage drives, internal algorithms and other Hadoop parameters. Table~\ref{table:dataset_properties} shows the data-set relevant features, selected by looking for those providing information into models, and their values.

\begin{table}[h!tbp]
\centering
\begin{tabular}{|l|l|}
	\hline
	\multicolumn{2}{|c|}{Benchmarks}\\ \hline
	\multicolumn{2}{|c|}{bayes, terasort, sort, wordcount, kmeans,}\\ 
	\multicolumn{2}{|c|}{pagerank, dfsioe\_read, dfsioe\_write }\\ \hline \hline
	\multicolumn{2}{|c|}{Hardware Configurations} \\ \hline
	Network & Ethernet, Infiniband \\ \hline
	Storage & SSD, HDD, Remote Disks \{1-3\} \\ \hline
	Cluster & \# Data nodes, VM description \\ \hline \hline
	\multicolumn{2}{|c|}{Software Configurations} \\ \hline
	Maps & 2 to 32 \\ \hline
	I/O Sort Factor & 1 to 100 \\ \hline
	I/O File Buffer & 1KB to 256KB \\ \hline
	Replicas & 1 to 3 \\ \hline
	Block Size & 32MB to 256MB \\ \hline
	Compression Algs. & None, BZIP2, ZLIB, Snappy \\ \hline
	Hadoop Info & Version \\ \hline
\end{tabular}
\vspace{-2 mm}
\caption{Configuration parameters on data-set}
\label{table:dataset_properties}
\end{table}

\begin{figure*}[tbp]
\centering
\small
\begin{tabular}{|l|l|l|l|l|l|l|l|l|l|}
\hline
id\_exec & id\_cl & bench & exe\_time & start\_time & end\_time & net & disk & bench\_type & maps\\
2 & 3 & terasort & 472.000 & 2014-08-27 13:43:22 & 2014-08-27 13:51:14 & ETH & HDD & HiBench & 8\\
\hline
\multicolumn{10}{c}{~} \\ \hline
iosf & replicas & iofilebuf & compression & blk\_size & \# data nodes & VM\_cores & VM\_ram & validated & version\\
10 & 1 & 65536 & None & 64 & 9 & 10 & 128 & 1 & 1\\
\hline
\end{tabular}
\caption{Example of logged observation on the data-set}
\label{figure:instance_example}
\end{figure*}

Figure~\ref{figure:instance_example} shows an example execution logged on the ALOJA framework. From here we distinguish the input data, introduced by the user executing the benchmark; the result of execution as the total time spent and time-stamps; and other key information that identifies the execution. We focus our interest on the elapsed time for a given execution as this can determine the cost of execution and indicate whether the execution is successful. Other important output data includes resources consumed such as usage statistics for CPU, memory, bandwidth and storage. We center our focus and efforts on learning and predicting the execution time for a given benchmark and configuration as our initial concern is to reduce the number and duration of executions required to characterize (learn) the system behavior. 

The collection of traces is part of the ALOJA framework~\cite{bsc:hadoop} and encourage the academic community and industry to make use of these results and/or to contribute to the corpus of results we have begun to generate.

\subsection{Methodology}

\begin{figure}[h!tbp]
\centering
\includegraphics[width=0.95\linewidth]{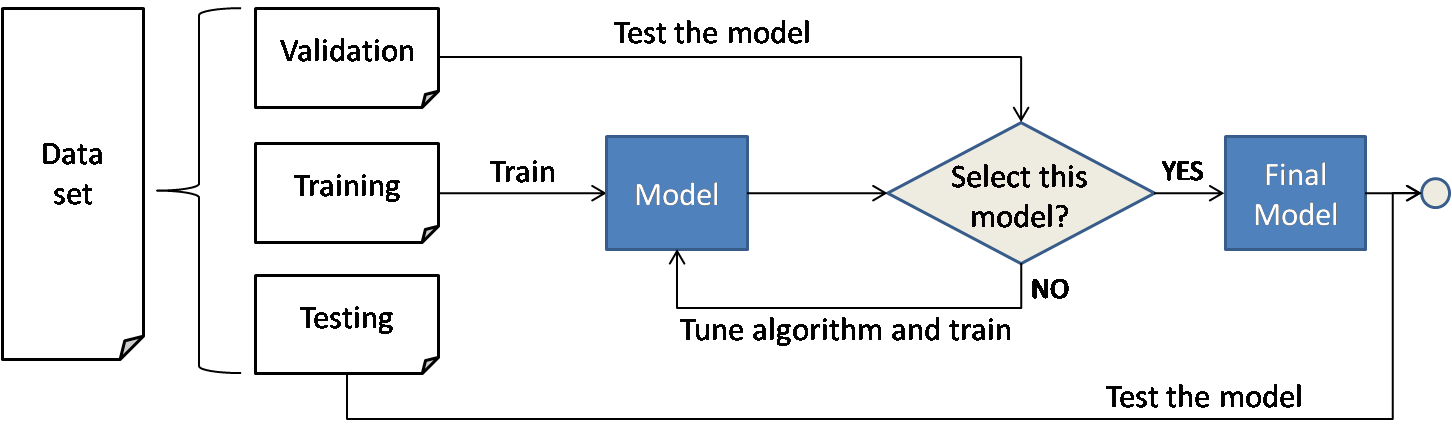}
\caption{Data-set splitting and learning schema}
\label{figure:data-splits-schema}
\end{figure}

The learning methodology is a 3-way step model involving training, validation and testing; see Figure~\ref{figure:data-splits-schema}. Data-sets in ALOJA are split (through random sample) and two subsets used to train and validate the obtained model. A selected algorithm (taken from those detailed in section~\ref{ssec:algorithms}) learns and characterizes the system, also identifying and retaining the 'best' parameters (testing them on the validation split) from a list of preselected input parameters. The third subset is used to test the best-from-parameters model. All learning algorithms are compared through the same test subset.

\begin{figure}[h!tbp]
\centering
\includegraphics[width=0.95\linewidth]{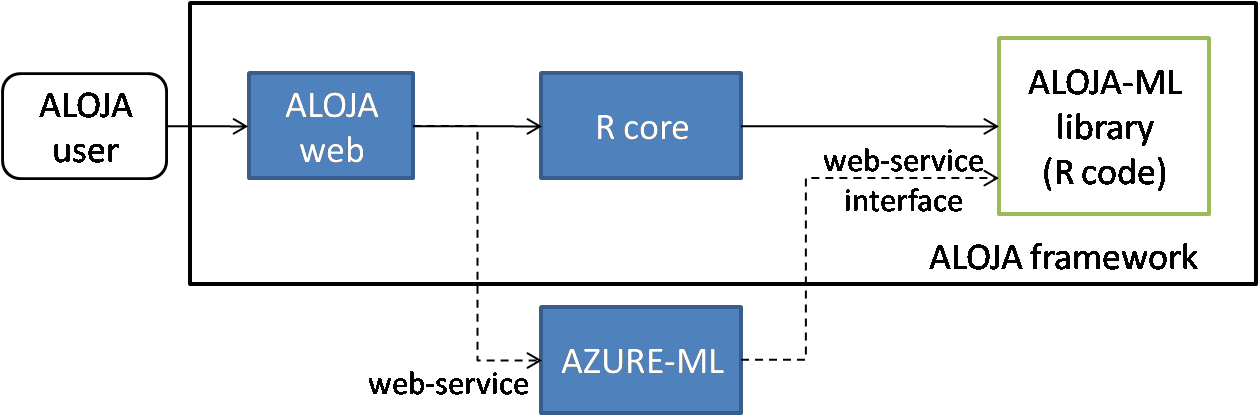}
\caption{Schema of AZURE-ML on the ALOJA framework}
\label{figure:alojaml-azureml}
\end{figure}

The tool-set is available as an R library in our on-line {\em github} repository\footnote{https://github.com/Aloja/aloja-ml}.  In our system the tools are called from the ALOJA web service but access can be initiated from any R-based analysis tool through R's ability to load remote functions from the Web. In this way any service or application can call our set of methods for predicting, clustering or treating Hadoop executions. In addition, our machine learning framework can be embedded on Microsoft Azure-ML services, delegating most of the learning process to the Cloud thereby reducing the ALOJA framework code footprint and enabling scaling the cloud.  The architecture can be seen in Figure~\ref{figure:alojaml-azureml}. The learning algorithms used here are part of the R {\em stats} and {\em nnet} packages, and RWeka~\cite{Hall:2009:WDM:1656274.1656278}.

\subsection{Algorithms}
\label{ssec:algorithms}

At this time, the ALOJA-ML user can choose among four learning methods: Regression Trees, Nearest Neighbors, Feed-Forward Artificial Neural Networks, and Polynomial Regression. Each of these has  different mechanisms of learning and strengths/weaknesses in terms of handling larges volume of data, being resilient to noise, and dealing with complexity.

Regression tree algorithm: we use the M5P~\cite{Quinlan1992,Wang1997} from the RWeka toolkit. The parameter selection (number of instances per branch) is selected automatically after comparing iteratively the prediction error of each model on the validation split.

Nearest neighbor algorithm: we use the IBk~\cite{Aha1991}, also from the RWeka toolkit. The number of neighbors is also chosen the same way as parameters on the regression trees.

Neural networks: we use a 1-hidden-layer FFANN from {\em nnet} R package~\cite{nnet:venables2002} with pre-tuned parameters as the complexity of parameter tuning in neural nets require enough error and retrial to not provide a proper usage of the rest of tools of the framework. Improving the usage of neural networks, including the introduction of deep learning techniques, is in the road-map of this project.

Polynomial regression: a baseline method for prediction, from the R core package~\cite{stats:rcore2014}. Experiences with the current data-sets have shown that linear regression and binomial regression do not produce good results, but trinomial approximates well. Higher degrees have been discarded because of the required computation time, also to prevent over-fitting.

\subsection{Execution Time Prediction}

Predicting the execution time for a given benchmark and configuration is the first application of the toolkit. Knowing the expected execution time for a set of possible experiments helps decide which new tasks to launch and their priority order (in case of time constraints or the need for immediate insights).

\begin{figure}[h!tbp]
\centering
\includegraphics[width=0.95\linewidth]{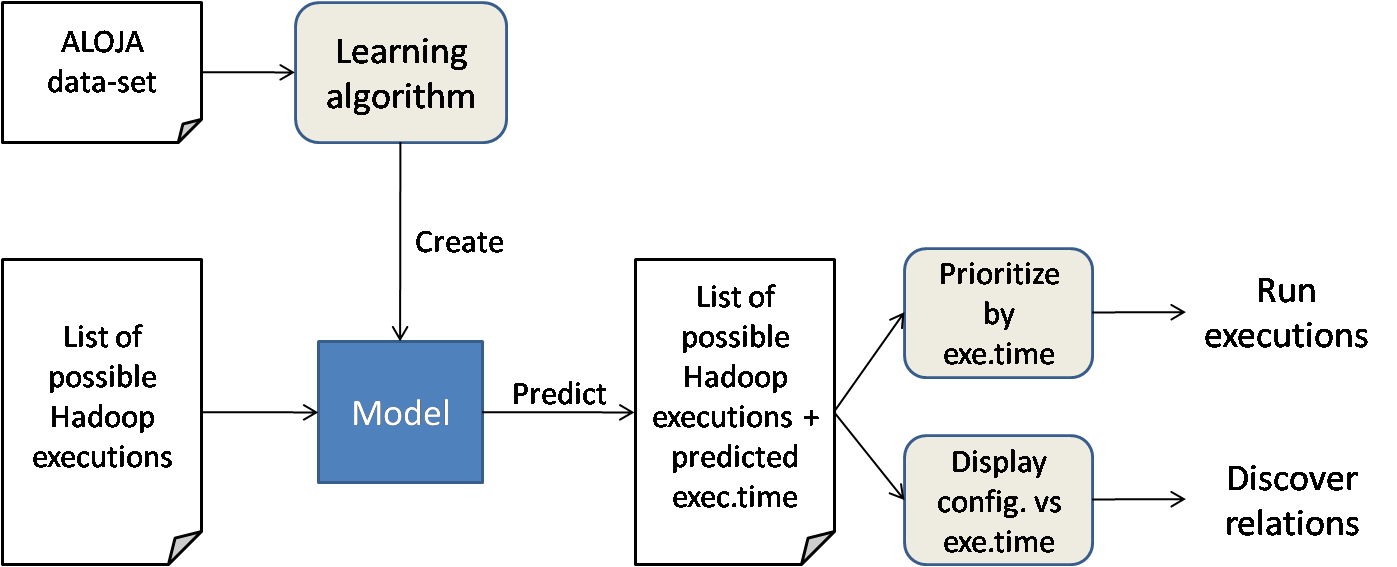}
\caption{Learning and prediction data-flow schema}
\label{figure:learn-schema}
\end{figure}

As said previously, the ALOJA data-set is used as training, validation and testing, with separate data splits blindly chosen. Deciding the best sizes for such splits becomes a challenge, as we seek to require as few benchmark executions as possible to train while maintaining good prediction accuracy. Also, at this time, we would like to execute as few jobs as possible to predict the rest of possible executions. Put simply, we look to build an accurate model from the minimum number of observations. As this specific problem is to explore in the next subsection~\ref{ssec:minconf}, here we check the accuracy of predictors given different sizes of training sets. Figure~\ref{figure:learn-schema} shows the learning and prediction data flow.

Furthermore, in this case we observe how much we can learn from logs and traces obtained from Hadoop executions, how much we can generalize when having different benchmark executions, also how ML improves prediction in front of other {\em rule of the thumb} techniques applied in the field.

\subsubsection{Validation and Comparisons}

As explained in subsection~\ref{ssec:algorithms}, we have several prediction algorithms from which to create our models, as well as different parameters and choices on the training type. Figure~\ref{table:learning-error} shows the average errors, absolute and relative, for the validation and the testing process for each learning algorithm. In a first exploration, we iterated through some pre-selected parameters and different split sizes for training, validation and testing data-sets. Figure~\ref{table:learning-error-splits} shows how the size of training versus validation affects the model accuracy.

\begin{figure*}[h!tbp]
\center
\begin{tabular}{| l | c | c | c | c | c |}
    \hline
    Algorithm & MAE Valid. & RAE Valid. & MAE Test & RAE Test & Best parameters \\ \hline \hline
	Regression Tree & 135.19523 & 0.16615 & 323.78544 & 0.18718 & M = 5 \\ \hline
	Nearest Neighbors & 169.64048 & 0.18968 & 232.01521 & 0.18478 &  K = 3 \\ \hline
	FFA Neural Nets & 189.60124 & 0.24541 & 333.93250 & 0.26099 & 5 neurons (1-hl), 1000 max-it, decay $5\cdot 10^{-4}$\\ \hline
	Polynomial Regression & 167.98270 & 0.2321720 & 354.93680 & 0.2541475 & degrees = 3\\ \hline
\end{tabular}
\caption{Mean and Relative Absolute Error per method, on best split and parameters found}
\label{table:learning-error}
\end{figure*}


\begin{figure*}[h!tbp]
\center
\begin{tabular}{| l | c | c | c |}
    \hline
    Algorithm & RAE 50/25/25 Split & RAE 37.5/37.5/25 Split & RAE 20/55/25 Split\\ \hline \hline
	Regression Tree & 0.18718 & 0.17903 & 0.22738 \\ \hline
	Nearest Neighbors & 0.18478 & 0.21019 & 0.30529 \\ \hline
	FFA Neural Nets & 0.27431 & 0.26099 & 0.27564 \\ \hline
	Polynomial Regression & 0.2541475 & 0.2602514 & 0.9005622 \\ \hline
\end{tabular}
\caption{RAE per method on test data-set, with different $\%$ splits for Training/Validation/Testing}
\label{table:learning-error-splits}
\end{figure*}

As can been seen in figure~\ref{table:learning-error-splits}, using regression trees and nearest neighbor techniques we can model and predict the execution time for our Hadoop traces. We consider that, with a more dedicated tuning, neural networks and deep believe learners could improve results. After testing, linear and polynomial regressions were set aside they achieve poor results when compared with the other algorithms and the time required to generate the model is impractical for the amount of data being analyzed.


Another key concern was whether a learned model could be generalized, using data from all the observed benchmarks, or would each execution/benchmark type require its own specific model. One motivation to create a single general model was to reduce the overall number of executions and to generate powerful understanding of all workloads. Also, there was an expectation that our selected algorithms would be capable of distinguishing the main differences among them (e.g., a regression tree can branch different trees for differently behaving benchmarks). On the other hand, we knew that different benchmarks can behave very differently and generalizing might compromise model accuracy. Figure~\ref{table:learning-general-specific} shows the results for passing each benchmarks test data-set through both a general model and a model created using only its type of observations.

\begin{figure}[h!tbp]
\center
\begin{tabular}{| l | c | c |}
    \hline
    Benchmark & RAE General Model & RAE Specific Model \\ \hline \hline
	bayes & 0.21257 & 0.16311 \\ \hline
	dfsio\_read & 0.45237 & 0.18472 \\ \hline
	dfsio\_write & 0.26468 & 0.16861 \\ \hline
	k-means & 0.22700 & 0.20065 \\ \hline
	pagerank & 0.76518 & 0.17862 \\ \hline
	sort & 0.42930 & 0.17535 \\ \hline
	terasort & 0.18190 & 0.19524 \\ \hline
	wordcount & 0.16282 & 0.17546 \\ \hline
\end{tabular}
\caption{Comparative for each benchmark, predicting them using the general vs. a fitted model with regression trees. The other algorithms show same trends}
\label{table:learning-general-specific}
\end{figure}

We are concerned about the importance of not over-fitting models, as we would like to use them for predicting unseen benchmarks similar to the ones already known in the future. Also, the fact that there are some benchmarks with more executions conditions the general model. After seeing the general vs specific results, we are inclined to use benchmark-specific models in the future, but not discarding using a general one when possible.

After the presented set of experiments and derived ones, not presented here for space limitations, we conclude that we can use ML predictors for not only predict execution times of unseen executions, but also for complementing other techniques of our interest, as we present in the following section. Those models provide us more accuracy that techniques used as {\em rules of thumb} like Least Squares or Linear Regressions (achieving LS with each attribute an average RAE of $2.0256$ and Linear Regression a RAE of $0.80451$).

\subsubsection{Applicability}

There are several uses on Hadoop scenarios for such prediction capability. One of them is predicting the performance of known benchmarks on a new computing cluster, as far as we are able to describe this new cluster. Having the hardware configuration of such cluster we can predict our benchmark executions with the desired software configuration. In case of a new benchmark entering the system, we can attempt to check if any of the existing models for specific benchmarks fits the new one. And then treat the new benchmark as the known one, or expand or train a new model for this benchmark.

Another important situation is filling values and configurations, or creating new $\langle$configuration, execution time$\rangle$ combinations for a benchmark, in order to observe the expected importance of a given parameter or a given range of values on a parameter, with reference to performance.

Further, an important application of such prediction is to know the size in time of a Hadoop workload, aiming to schedule properly the execution inside a cluster. Being able to find the proper combination of parameters for each workload also their allocation in time and placement becomes an interesting problem. Including predictive capabilities into a job scheduling problem can improve consolidation and de-consolidation processes~\cite{DBLP:conf/icpp/BerralGT13}, and so reduce resource consumptions by maintaining any quality of service or job deadline preservation.

Finally, the next section presents another specific uses for the predicting model, with high priority in the road-map of the ALOJA project. These are 1) anomaly detection, by detecting faulty executions through comparing their execution times against the predicted ones; and 2) identification of which executions would best model a given Hadoop scenario or benchmark, by clustering the execution observations, and taking the resulting cluster centers as recommended configurations. Next section focuses on these two cases in detail.

\section{Selected Use Cases}
\label{sec:experiences}

This section exposes two types of experiences on the use of ALOJA with machine learning tools, and some discoveries obtained over the Hadoop data-set.

\subsection{Anomaly Detection}

Detecting which executions of our data-sets can be considered valid, flagged for revision or directly discarded, is the second application of the prediction models. Detecting automatically executions susceptible of being failures, or even executions not modeled properly, can save time to users who must check each execution, or can require less human intervention on setting up {\em rules of thumb} to decide which executions to discard. 

\begin{figure}[h!tbp]
\centering
\includegraphics[width=0.95\linewidth]{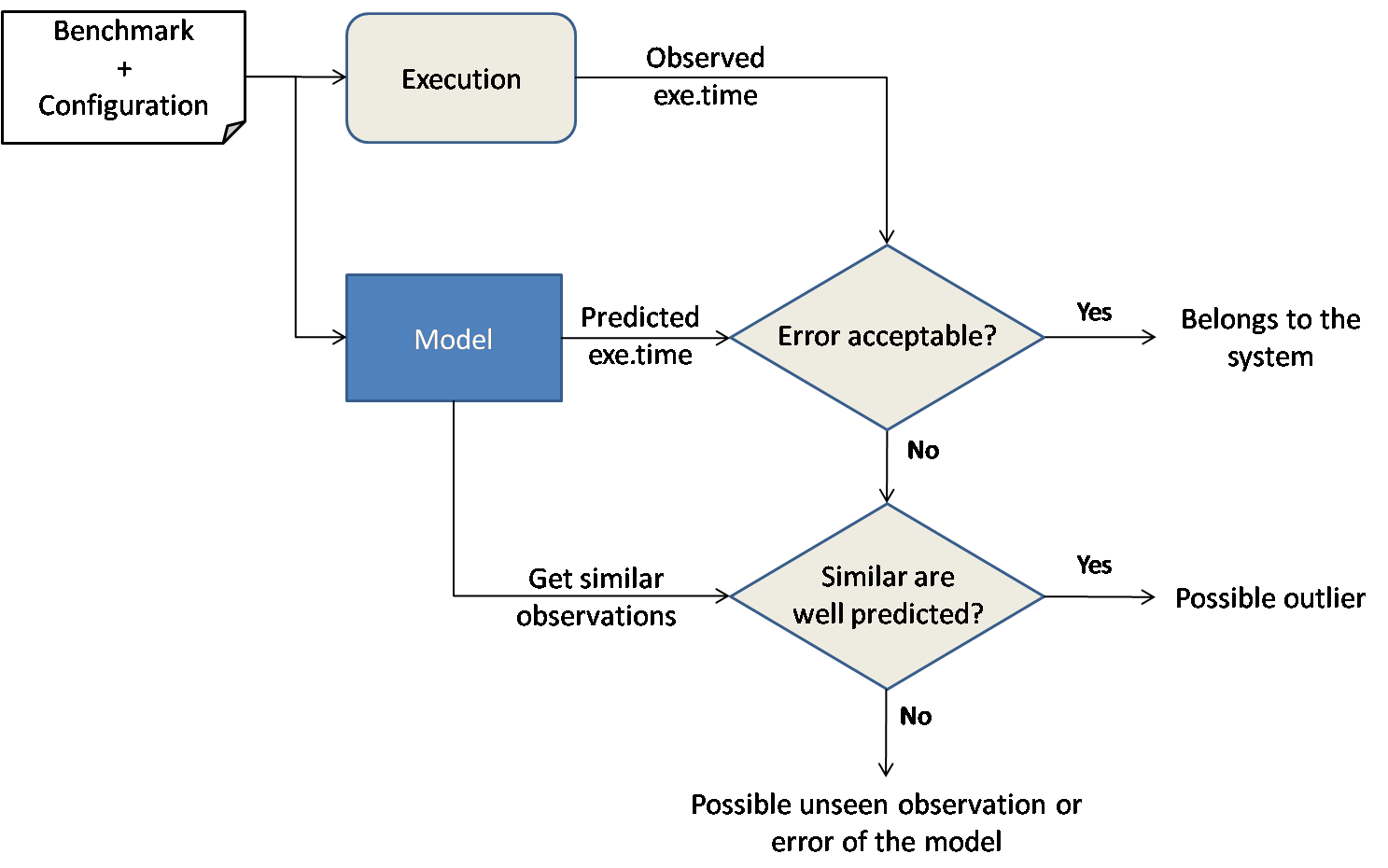}
\caption{Anomaly detection mechanism schema}
\label{figure:outlier-schema}
\end{figure}

With Hadoop workload behaviors modeled, model-based anomaly detection methods are applied. Considering the learned model as the `rule that explains the system', any observation that does not fit into the model (this is, the difference between the observed value and the predicted value is bigger than expected), is flagged as anomalous. Here we flag anomalous data-set entries as {\em warnings} and {\em outliers}. A {\em warning} is an observation whose error respect to the model is
$n$ standard deviations from the mean.
An {\em outlier} is a mispredicted observation where other similar observations are well predicted. I.e an outlier is a warning that, for all its neighbor observations (those ones that differ in less than $h$ attributes, or with Hamming distance $<h$), more than a half are well predicted by the model. Figure~\ref{figure:outlier-schema} shows the anomaly decision making schema.

In this case we observe how our automatic method can detect outliers or anomalous executions, against a human review and a human-based rules review.

\subsubsection{Validation and Comparisons}




Once we can model the ALOJA data-set, and knowing that it can contain anomalous executions, we proceed to auto-analyze itself with the anomaly detection method. Here we perform two kind of experiments, one testing the data of a single benchmark with a model learned from all the observations, and one testing it against a model created from its specific kind of observations. We select the observations belonging to the {\em Terasort} benchmark, with 7844 executions, and the regression trees algorithm from the previous subsection.

After applying our method, using as parameters $h=\{0...3\}$, $n=3$, and a model learned from all the observed executions, we detect 20 executions of Terasort with a time that does not match with the expected execution time. After reviewing them manually, we detect that those executions are valid executions, meaning that they finished correctly, but something altered their execution, as other repetitions finished correctly and in time. Further, when learning from Terasort observations only ($7800$), the model is more fitted and is able to detect 4 more executions as anomalous, which in the general model where accepted because of similarities with other benchmark executions with similar times. From here on we recommend to validate outlier executions from models trained exclusively with similar-kind executions.

\begin{figure}[h!tbp]
\centering
\includegraphics[width=0.95\linewidth]{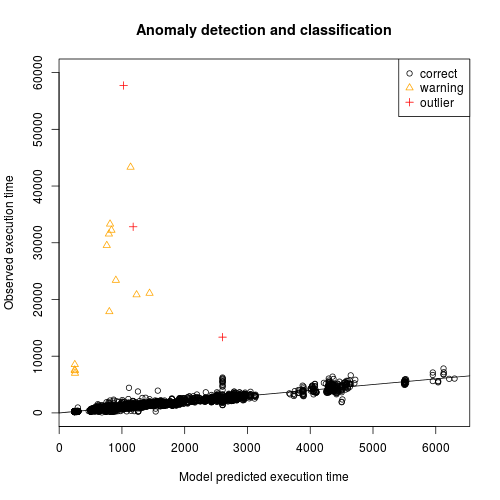}
\vspace{-2mm}
\caption{Automatic outlier detection ($h=0$, $n=3$)}
\label{figure:outlier-results}
\end{figure}

Testing with different values of Hamming distance shows low variation, as outliers are easily spotted by its error, also confirmed by several neighbors at distance 1. Setting up distance 0, where an observation only is an outlier when there are at least two identical instances with an acceptable time, marks 17:24 observations as warnings, and 7:24 as outliers. Such warnings can be set to revision by a human referee to decide whether are outliers or not. Figure~\ref{figure:outlier-results} show the comparative of observed versus predicted execution times, marking outliers and warnings.

Notice that the confusion matrix shows anomalies considered legitimate by the automatic method. After human analysis, we discovered that such ones are failed executions with a very low execution time, whose error at prediction is also low, so the method does not detect them. This let us see that for such detection, a manual rule can be set, as if an execution does not exceed a minute, belongs to a failed execution.

\begin{figure}[h!tbp]
\centering
\vspace{-3mm}
\includegraphics[width=0.95\linewidth]{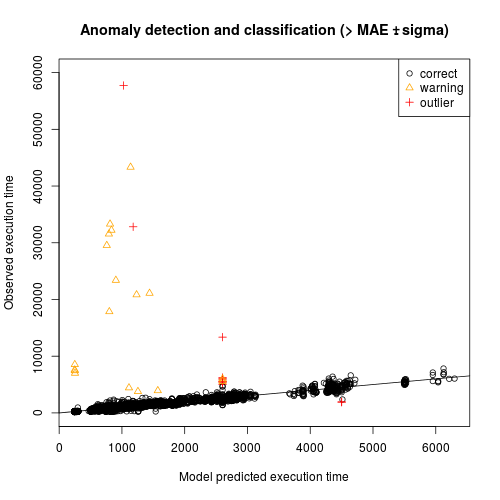}
\vspace{-2mm}
\caption{Automatic outlier detection ($h=0$, $n=1$)}
\label{figure:outlier-results2}
\end{figure}

Also, we may decide to reduce the number of directly accepted observations by lowering $n$ (standard deviations from the mean) from 3 to 1, putting under examination from a $1\%$ of our data-set up to a $34\%$. In this situation, we increase slightly the number of detections to 38 (22 warnings and 16 outliers). Figure~\ref{figure:outlier-results2} show the comparative of observed versus predicted execution times, marking outliers and warnings. Also figure~\ref{figure:outlier-confmatrix} shows the confusion matrices for the automatic versus a manual outlier tagging, automatic versus {\em rule of thumb} (semi-automatic method where ALOJA checks if the benchmark has processed all its data), and automatic warnings versus manual classification as incorrect or ``to check'' (which is if the operator suspects that the execution has lasted more than 2x the average of its similar executions).

\begin{figure}[h!tb]
\center
\begin{subfigure}{0.5\linewidth}
\small
\begin{tabular}{ r | c | c | }
  automatic $\rightarrow$ & & \\ manual $\downarrow$ & Outlier & OK \\ \hline
  Anomaly & 12 & 22 \\ \hline
  Legitimate & 4 & 7786 \\ \hline
\end{tabular}
\end{subfigure}%
~~~
\begin{subfigure}{0.5\linewidth}
\small
\begin{tabular}{ r | c | c | }
  automatic $\rightarrow$ & & \\ {\em semi-auto.} $\downarrow$ & Outlier & OK \\ \hline
  Anomaly & 7 & 0 \\ \hline
  Legitimate & 9 & 7786 \\ \hline
\end{tabular}
\end{subfigure}
\\
\begin{subfigure}{0.5\linewidth}
\small
\begin{tabular}{ r | c | c | }
  automatic $\rightarrow$ & & \\ manual $\downarrow$ & Warning & OK \\ \hline
  {\em to check} & 22 & 0 \\ \hline
  Legitimate & 0 & 7786 \\ \hline
\end{tabular}
\end{subfigure}
\vspace{-2mm}
\caption{Confusion matrices for different methods}
\vspace{-2mm}
\label{figure:outlier-confmatrix}
\end{figure}


Finally, besides using this method to test new observations, we can re-train our models discarding the observations considered inaccurate. In the first regression tree case, shown in the previous subsection, by subtracting the observations marked as outlier we are able to go from a prediction error of $0.16615$ to $0.10873$ on validation, and $0.18718$ to $0.11912$ with the test data-set.

\subsubsection{Use cases}

Spotting failed executions in an automatic way saves time to users, but also let the administrators know when elements of the system are wrong, faulty, or have unexpectedly changed. Further, sets of failed executions with common configuration parameters indicate that it is not a proper configuration for such benchmark; or failing when using specific hardware shows that such hardware should be avoided for those executions.

Furthermore, highlighting anomalous executions is always positive for easing analysis, even more when having more than 16000 executions (and the +300.000 other system performance traces that come with each execution trace). Also it allows to use other learning models less resilient to noise.

\subsection{Guided Benchmarking}
\label{ssec:minconf}

When modeling a benchmark, a set of configurations, or a new hardware set up, some executions must be performed to observe its new behavior. But executions cost money and time, so we want to run as few sample executions as possible. This is, run the minimum set of executions that define the system with enough accuracy.

We have a full data-set of executions, and we can get a model from it. From it we can attempt to obtain which of those executions are the most suitable to run on a new system or benchmark, and use the results to model it; or we can use the data-set and model to obtain which executions, seen or unseen in our data-set, can be run and used to model.
The data-set, as obtained from random or serial executions, can contain similar executions, introduce redundancy or noise. And find which minimal set of executions are the ones that minimize the amount of training data is a combinatorial problem on a big data-set.

\begin{figure}[h!tbp]
\centering
\includegraphics[width=0.95\linewidth]{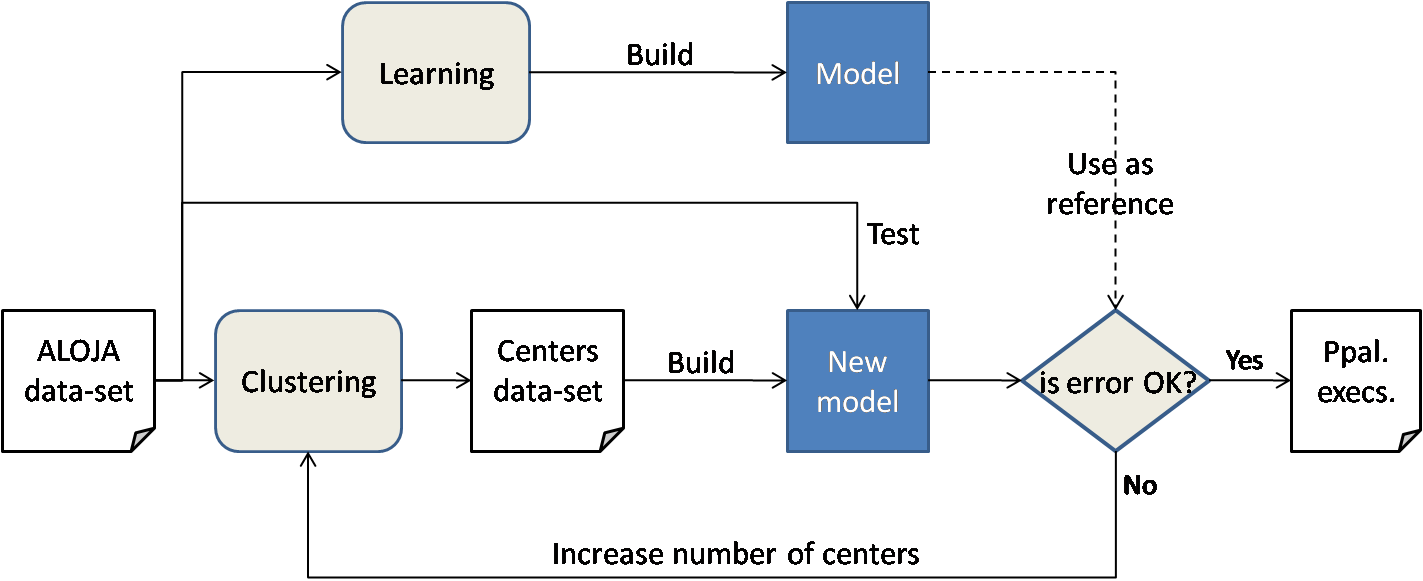}
\vspace{-1mm}
\caption{Finding recommended executions schema}
\vspace{-2mm}
\label{figure:minconfs-schema}
\end{figure}

Our proposal is to cluster our observed executions (i.e., apply the {\em k-means} algorithm~\cite{stats:rcore2014}), obtain for each cluster a representative observation (i.e., its centroid), and use the representatives as the recommended set of executions. Determine the number of clusters (recommendations) required to cover most of the information is the main challenge here. At this time we iterate through a range of $k$, as figure~\ref{figure:minconfs-schema} displays, reconstructing a the model with those recommended observations, and testing it against our data-set or against a reference model. From here on, we decide when the error is low enough or when we exceed the number of desired executions.

\subsubsection{Validation and Comparisons}

\begin{figure*}[h!tbp]
\centering
\vspace{-3mm}
\includegraphics[width=0.95\linewidth]{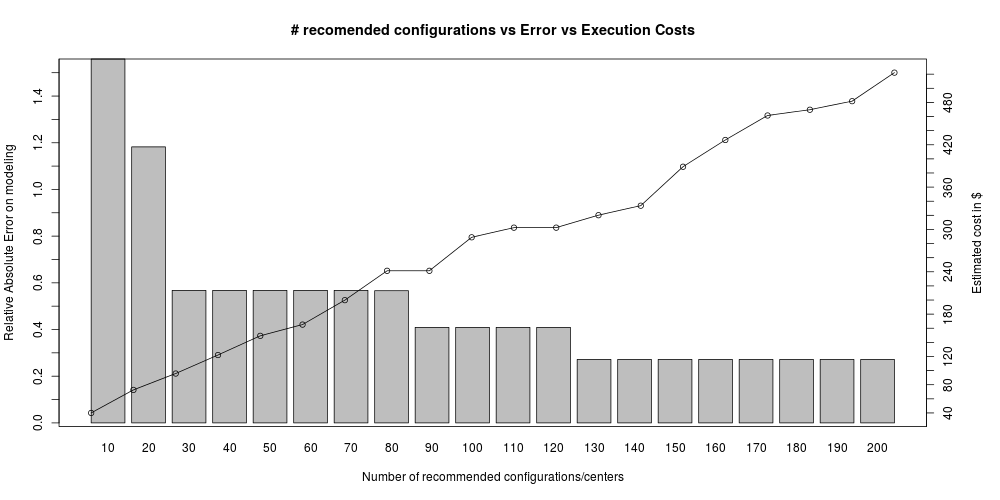}
\vspace{-4mm}
\caption{Number of recommended executions vs. error in modeling vs. execution cost}
\vspace{-4mm}
\label{figure:minconfs-k-error-cost}
\end{figure*}

For each iteration, looking for $k$ clusters, we compute the error of the resulting model against the reference data-set or model. Also we can estimate the cost of running those executions, as we have the it also we know the average execution cost of $6.85~\$/hour$ for the clusters we tested. Notice that the estimated execution time is from the seen data-set, and applying those configurations on new clusters or unseen components may increase or decrease the values of such estimations, and it should be treated as a guide more than a strict value. Figure~\ref{figure:minconfs-k-error-cost} shows the evolution of the error and the execution cost per each $k$ recommendations from our ALOJA data-set. As we expected, more executions implies more accuracy on modeling and predicting, but more cost and execution time. 

Further, to test the method against a new cluster addition, we prepared new a setup (on premise, 8 data nodes, 12 core, 64 RAM, 1 disk), and run some of the recommendations obtained from our current ALOJA data-set. We get 6 groups of recommendations with from $k=\{10...60, step=10\}$, and we executed them in order (first the group of $k=10$ and so on, removing in this case the repeated just to save experimentation time). We found that with only those 150 recommendations we are able to learn a model with good enough accuracy (tested with all the observed executions of the new cluster), compared to the number of executions needed from the ALOJA data-set to learn with similar accuracy.

\begin{figure}[h!tbp]
\centering
\vspace{-3mm}
\includegraphics[width=0.85\linewidth]{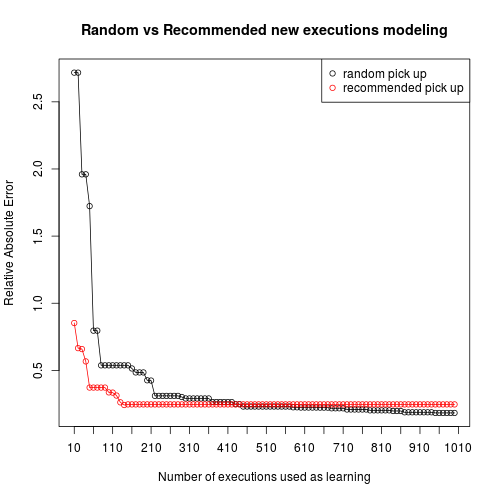}
\vspace{-4mm}
\caption{Random vs. recommended executions}
\vspace{-3mm}
\label{figure:minconfs-learning-rate}
\end{figure}

Figure~\ref{figure:minconfs-learning-rate} shows the comparative of learning a model with $n$ random observations picked from the ALOJA data-set, seeing how introducing new instances to the selected set improves the model accuracy, against picking the new executed instances (this case in order of recommendation groups), and see how it improves the learning rate of the new cluster data-set. We can achieve low prediction errors very quickly, in front of a random execution selection.

\subsubsection{Use cases}

Often executions on a computing cluster are not for free, or the amount of possible configuration (HW and SW) to test are huge. Finding the minimal set of executions to be able to define the behavior of our Hadoop environment helps to save time and/or money. Also, an administrator or architect would like to prioritize executions, running first those that provide more information about the system, and then run the rest in descending order of relevance. This is useful when testing or comparing our environment after modifications, sanity checks, or validating clone deployments.

Further, it is usual when adding new benchmarks or resources, that the new benchmark is similar in behavior to another previously seen, or that a hardware component is similar in behavior to another. Instead of testing it from random executions, we could use the principal executions for the most similar seen environments to test it, and although results can not fit well with previous models (in fact the new environment can be different), use the new observed results as a starting point to create a new model. The study of how it performs well against other example selection methods for Hadoop platforms and brand new benchmarks is in the ALOJA-ML road-map for near future research.

\section{Summary and Conclusions}
\label{sec:conclusions}

In this article we described ALOJA-ML, a rich tool-set for carrying on automated modeling and prediction tasks over benchmarking data repositories, with particular focus on the BigData domain. The goal of ALOJA-ML is to assist performance engineers, infrastructure designers, data scientists writing BigData applications and cloud providers in minimizing the time required to gain knowledge about deployment variables for BigData workloads. ALOJA-ML takes large performance data repositories available in the ALOJA project and through several Machine Learning techniques identifies key performance properties of the workloads. This information can be later leveraged to predict performance properties for a workload execution on a given set of deployment parameters that have not been explored before in the testing infrastructure. In particular, in this paper we take the case of the Hadoop ecosystem, which is the most widely adopted BigData processing framework currently available, to describe the approach used in ALOJA-ML.

The article describes: a) the methodology followed to process and model the performance date; b) the group of learning algorithms selected for the experiments; c) and the characteristics of the public data set (consisting of more than 16,000 detailed execution logs at the time of the experiments but continuously growing) used for training and validation if the learning algorithms. Through our experiments, we exposed and demonstrated that using our techniques we are able to model and predict Hadoop execution times for given configurations, with a small relative error around $0.20$ depending on the executed benchmark. Further we passed out data-set through an automated anomaly detection method, based on our obtained models, with high accuracy respect a manual revision. Also we deployed a new Hadoop cluster, running the recommended executions from our method, and tested the capability of characterizing it with little executions; finding that we can model the new deployment with fewer executions than by randomly selecting executions.

The work presented includes two selected use cases of the ALOJA-ML tool-set in the scope of the ALOJA framework. The fist one is the use of the learning tools for guiding the costly task running experiments for the online performance data-set. In this case, ALOJA-ML assists the framework at the time of selecting the most representative runs of an application that wants to be characterized in a given set of deployment, with the consequent reduction in terms of time and costs to produce relevant executions. The second use case assists in identifying anomalies on large sets of performance testing experiments, which is a common problem of benchmarking frameworks, and that needs to be addressed to get unbiased conclusions on any performance evaluation experiment.


\section{Future Work}
\label{sec:future}

The current road-map of ALOJA-ML includes to add new features, tools and techniques, improving the ALOJA framework for Hadoop data analysis and knowledge discovery. Learn how to deal with big amounts of data from Hadoop executions is among our principal interests. This knowledge and capabilities can be used to improve the comprehension and management of such platforms. We are open to new techniques or new kind of applications.\looseness=-1

Our next steps are:
1) study techniques to characterize computation clusters and benchmarks, to prepare Hadoop deployments to deal with seen and unseen workloads;
2) introduce new input and output variables, looking new sources of information from the system, and predicting other performance indicators like resource usage;
3) study in detail compatibilities and relations among configuration and hardware attributes, in order to detect impossible setups or executions incompatible with a given deployment;
4) improve the methods to select features, examples and learning parameters. Also introduce new learning algorithms and methodologies like deep-belief networks;
5) and add new executions to the ALOJA data-sets, with new deployments and benchmarks.

\section{Acknowledgments}
{\small This project has received funding from the European Research Council (ERC) under the European Union's Horizon 2020 research and innovation programme (grant agreement No 639595). This work is partially supported by the Ministry of Economy of Spain under contracts TIN2012-34557 and 2014SGR1051.}

\bibliographystyle{abbrv}
\bibliography{sigkdd15} 
\end{document}